\documentclass{article}

\usepackage{microtype}
\usepackage{graphicx}
\usepackage{multirow}
\usepackage{placeins}
\usepackage{booktabs} % for professional tables

\usepackage{hyperref}

\usepackage[preprint]{icml2026}

\usepackage{amsmath}
\usepackage{amssymb}
\usepackage{mathtools}
\usepackage{amsthm}

\usepackage[capitalize,noabbrev]{cleveref}

\theoremstyle{plain}

\theoremstyle{definition}

\theoremstyle{remark}

\usepackage[textsize=tiny]{todonotes}

\icmltitlerunning{Do Language Models Encode Knowledge of Linguistic Constraint Violations?}

\begin{document}

\twocolumn[
  \icmltitle{Do Language Models Encode Knowledge of Linguistic Constraint Violations?}



  \icmlsetsymbol{equal}{*}

  \begin{icmlauthorlist}
    \icmlauthor{Hardy}{equal,a,b}
    \icmlauthor{Sebastian Pad{\'o}}{a}

  \end{icmlauthorlist}

  \icmlaffiliation{a}{IMS, University of Stuttgart, Stuttgart, Germany}
  \icmlaffiliation{b}{Universitas Mikroskil, Medan, Indonesia}

  \icmlcorrespondingauthor{Hardy}{hardy@ims.uni-stuttgart.de}
  \icmlcorrespondingauthor{Sebastian Pad{\'o}}{pado@ims.uni-stuttgart.de}

  \icmlkeywords{Machine Learning, ICML}

  \vskip 0.3in
]



\printAffiliationsAndNotice{}  % no special notice (required even if empty)

\begin{abstract}
Large Language Models (LLMs) achieve strong linguistic performance, yet their internal mechanisms for producing these predictions remain unclear. We investigate the hypothesis that LLMs encode representations of linguistic constraint violations within their parameters, which are selectively activated when processing ungrammatical sentences. To test this, we use sparse autoencoders to decompose polysemantic activations into sparse, monosemantic features and recover candidates for violation-related features.

We introduce a sensitivity score for identifying features that are preferentially activated on constraint-violated versus well-formed inputs, enabling unsupervised detection of potential violation-specific features. We further propose a conjunctive falsification framework with three criteria evaluated jointly.

Overall, the results are negative in two respects: (1) the falsification criteria are not jointly satisfied across linguistic phenomena, and (2) no features are consistently shared across all categories. While some phenomena show partial evidence of selective causal structure, the overall pattern provides limited support for a unified set of grammatical violation detectors in current LMs.
  
\end{abstract}

\section{Introduction}

Large language models \citep[LLMs;][]{Radford2018ImprovingLU, brown2020language} have demonstrated strong performance on a broad range of linguistic benchmarks \citep[see][for a survey]{guoEvaluatingLargeLanguage2023}. A key area of success is their ability to learn and generalize grammatical regularities from raw text alone \citep{petersDissecting2018, zhang_unveiling_2024, tang_language-specific_2024}, suggesting that LLMs internally develop representations that are consistent with the structure of natural language. However, the presence of these regularities does not explain the mechanisms the model uses to distinguish well-formed sentences from their ill-formed counterparts. Because current research largely overlooks this mechanism, exactly how LLMs process grammar violation internally remains an open question. In this study, we analyze internal activations when computing representations for well-and ill-formed sentences. We investigate whether LLMs, specifically Transformer-based architectures \citep{vaswani2017attention}, encode specific `detector features' that activate upon encountering grammatical violations, thereby penalizing the sentence's log-likelihood. 

We ground our study in the theoretical framework of string probability and grammaticality proposed by \citet{hu2026can}. Under this framework, the probability of a string is governed by two latent variables: the message ($m$), representing the intended semantic content, and grammaticality ($G$), a binary variable indicating adherence to formal linguistic rules. Our search for `detector features' is essentially an inquiry into the mechanistic realization of this latent variable $G$ -- identifying where and how this theoretical constraint is instantiated within the model's architecture. 

For the isolation of these `detector features,' we utilize the contrastive structure of the BLiMP benchmark \citep{warstadtBLiMP2020} which pairs well- with one ill-formed sentences that differ in exactly one word. By employing these minimal pairs, we hold the message variable $m$ constant while varying the grammatical constraint. This approach allows for a controlled observation of the model’s internal response to isolated grammatical violations. However, model activations are often polysemantic and entangled \citep{elhage2022toy, bricken2023monosemantic}, which makes it difficult to isolate $G$ from the broader representation of the message $m$.

To address this, we employ Sparse Autoencoders \citep[SAEs;][]{Cunningham2024sparse, rajamanoharan2024jumpingaheadimprovingreconstruction, gao2025scaling}, which offer a finer-grained decomposition of the representation. By learning an overcomplete, sparse basis for a model's residual stream, SAEs produce individual features that are monosemantic by design. We utilize these SAEs to decompose the internal activations of LLMs as they process BLiMP's contrastive pairs.

Our key contributions are: (1) A \textbf{sensitivity score} $\Phi_j$ that ranks SAE features by their preferential activation on constraint-violating over well-formed inputs, enabling targeted identification of violation-specific features without supervision; (2) A \textbf{conjunctive falsification framework} with three testable criteria: causal effect on violations ($F_1$), stability of non-violations ($F_2$), and preferential effect on violations ($F_3$); (3) \textbf{Empirical results} across 13 BLiMP linguistic phenomena and six models show that $F_1$ passes broadly, $F_2$ and $F_3$ are the main bottlenecks, as violation-related features often  encode general fluency information.

\section{Related Work}
A significant body of work demonstrates that deep bidirectional language models \citep{devlin2019bertpretrainingdeepbidirectional} learn a rich hierarchy of contextual information that mirrors the classical NLP pipeline \citep{petersDissecting2018}. Lower layers specialize in word morphology and local syntax, while higher layers represent longer-range semantic relationships, such as pronominal coreference. Observational evidence suggests models can dynamically adjust this pipeline, revising lower-level decisions (e.g., entity typing) based on higher-level disambiguating information \citep{li2021bert}. This hierarchy is further reflected in anomaly detection: morphosyntactic violations typically trigger surprisal in earlier layers than semantic violations, while commonsense anomalies often produce no significant surprisal at any intermediate layer.

Recent mechanistic interpretability work, such as LinguaLens \citep{jing2025lingualens}, uses Sparse Autoencoders (SAEs) to decompose hidden states into monosemantic features spanning morphology, syntax, and pragmatics. Our method also uses SAEs for causal feature intervention, but with a narrower goal: We focus on the latent variable $G$ (grammaticality) defined by \citet{hu2026can}, enabling a targeted mechanistic study of grammatical constraint violations in model representations.

\section{Methodology}
\subsection{Dataset}
Let $\mathcal{D} = \{(x^{(i)}_{\text{good}},\, x^{(i)}_{\text{bad}})\}_{i=1}^{N}$ be a dataset of minimal pairs, where $x_{\text{good}}$ adheres to a linguistic constraint, and $x_{\text{bad}}$ violates it. We use BLiMP \cite{warstadtBLiMP2020}, which covers 13 linguistic phenomena. Within each pair, sentences differ minimally so that any preference difference isolates sensitivity to the targeted constraint rather than surface form. We split the BLiMP dataset 80:20 into a train (validation) and a test set, stratified by phenomenon. We use the validation set to tune hyperparameters such as top-$k$.
 
\subsection{Models and SAEs}
We experiment with five Transformer models spanning three families and a range of scales: GPT-2 small \citep{Radford2018ImprovingLU} (124M parameters), Gemma-3 \citep{gemmateam2025gemma3technicalreport} (270M, 1B, and 4B), and Qwen3.5 \citep{qwen35blog} (2B and 9B). For each model, we use the corresponding pre-trained Sparse Autoencoder (SAE) from SAELens \citep{bloom2024saetrainingcodebase}, Qwen Scope \citep{qwen_scope}, and Gemma Scope 2 \citep{mcdougall2025gemmascope2}, applied to the residual stream at layers $l$.  %\citep{gao2025scaling, %bussmann2025learningmultilevelfeaturesmatryoshka}.
We also follow Gemma Scope 2 by selecting four SAE layers at 25\%, 50\%, 65\%, and 85\% of model depth for all models.

An SAE encodes a dense residual stream activation $\mathbf{h}^{(l)}_t \in \mathbb{R}^D$ at layer $l$ and token position $t$ into a sparse feature vector $\mathbf{z}^{(l)}_t \in \mathbb{R}^k$ via a learned encoder and reconstructs it via a linear decoder. The active feature index set is a set of feature indexes that are activated:
\begin{equation}
A_{i,t}^{(l)}=\{j|z_{i,t,j}^{(l)}>0\}
\end{equation}
The SAEs differ in architecture across families: Gemma-3 uses Matryoshka SAEs \citep{bussmann2025learningmultilevelfeaturesmatryoshka}, while Qwen3.5 and GPT-2 use Top-K SAEs \citep{gao2025scaling}. We refer the reader to the  technical reports for details.

\subsection{Language Models}
Let $f(\theta)$ be the language model. The log likelihood of a sentence $x^{(i)} = (x_1^{(i)}, x_2^{(i)}, \dots, x_T^{(i)})$ is:
\begin{equation}
\mathcal{L}(x^{(i)}) = \text{log}\ P(x^{(i)};\theta)=\sum_{t=1}^T \text{log}\ P(x^{(i)}_t|x^{(i)}_{<t};\theta)
\end{equation}
For each pair $(x^{(i)}{\text{good}}, x^{(i)}{\text{bad}})$, we define the preference margin under both the original and ablated model as:
\begin{align}
M_i &= \mathcal{L}(x^{(i)}_{\text{good}}) - \mathcal{L}(x^{(i)}_{\text{bad}})\\
M_{i,l} &= \mathcal{L}(x^{\prime(i,l)}_{\text{good}}) - \mathcal{L}(x^{\prime(i,l)}_{\text{bad}}) 
\end{align}
where $M_i$ is the baseline margin and $M_{i,l}$ is the margin after ablating features in layer $l$. The sign of each margin indicates language modeling preference.

\subsection{Sensitivity Score-based Candidate Feature Set}

The \emph{sensitivity score} for a feature $j$ at a layer $l$ is:
\begin{equation}
    \Phi^{(l)}_j \;=\;
    \frac{1}{|\mathcal{D}|\cdot T}
    \sum_{i=1}^{|\mathcal{D}|}\sum_{t=1}^{T}
    \!\Bigl(z^{(l)}_{i,t,j,\text{bad}} - z^{(l)}_{i,t,j,\text{good}}\Bigr)
    \label{eq:phi}
\end{equation}
A positive value $\Phi^{(l)}_j$ indicates that the feature $j$ fires more strongly on constraint-violating sentences; a negative value indicates greater activation on well-formed sentences. Thus, features with a (strong) positive $\Phi$ are candidates for constraint violation indicators. We select the top-$k$ candidate features by descending $\Phi^{(l)}_j$ as the candidate violation feature set $\hat{A}^{(l)}$.
 
\subsection{Intervention through Ablation}
\label{sec:falsification}

Our general approach is to assess the causal role of our feature candidates by removing (ablating) them from the 
representations and compare the non-ablated and ablated versions. 
Concretely, for each layer and sentence, we run three forward passes through the SAE-augmented model: (1) \textbf{Baseline.} The SAE is active, but no features are modified; (2) \textbf{Targeted ablation.} All features within the candidate set $\hat{A}^{(l)}$ are set to zero, suppressing the identified violation signal; (3) \textbf{Random control.} An equal-sized random feature set $\hat{R}^{(l)}$, drawn from non-zero-$\Phi$ features excluding $\hat{A}^{(l)}$, is ablated instead. We use the superscript $M^\text{ctrl}$ to denote the random control preference margin.
 
We define three \emph{preference margins} capturing the change in the sum of log-likelihood relative to baseline:
% \marginpar{$\mathcal L$ with *two* superscripts not introduced!}
\begin{align}
    M_{i,l,\text{bad}}  &= \mathcal{L}(x'^{(i,l)}_{\text{bad}})
                            - \mathcal{L}(x^{(i,l)}_{\text{bad}})   \label{eq:mcv}\\
    M_{i,l,\text{good}} &= \mathcal{L}(x'^{(i,l)}_{\text{good}})
                            - \mathcal{L}(x^{(i,l)}_{\text{good}})  \label{eq:mwf} \\
    M_{i,l,\text{pref}} &= \mathcal{L}(x'^{(i,l)}_{\text{bad}})
                            -  \mathcal{L}(x'^{(i,l)}_{\text{good}})  \label{eq:msel} 
\end{align}

%\subsection{Falsification Criteria}

We can now formulate three specific hypotheses regarding these preference margins that we all expect to be true if the candidate features in $\hat A$ indeed indicate constraint violation. This makes our experiment formally a conjunctive falsification test. 
 
\begin{description}
    \item[$F_1$: Causal effect on violations.] The targeted ablation margins on constraint-violation sentences should be positive and significantly larger than those from random controls (one-sided Wilcoxon signed-rank test, $p \ge \alpha$). This serves as a basic sanity check that ablation increases the likelihood of constraint-violating sentences.
    \begin{equation}
    \bar{M}_{l,\text{bad}} = \frac{1}{|\mathcal{D}|}\sum_i M_{i,l,\text{bad}}, \ \text{with} \ \bar{M}_{l,\text{bad}} > 0
    \end{equation}
    %, $p < \alpha$.
    
    \item[$F_2$: Stability of non-violations.] The targeted ablation margins on well-formed sentences remain within a small threshold and are not significantly different from the random control margins (two-sided Wilcoxon signed-rank test, $p \ge \alpha$). This rules out non-specific fluency degradation: the ablated features must affect ungrammatical sentences without impacting grammatical ones.
    \begin{equation}
    \bar{M}_{l,\text{good}} = \frac{1}{|\mathcal{D}|}\sum_i M_{i,l,\text{good}}, \ \text{with} \ \left|\bar{M}_{l,\text{good}}\right| < 0.05
    \end{equation}
    A stricter threshold, such as $0.01$ would be preferable under ideal conditions. However, we allow a small tolerance to account for the inherent noise introduced by activation ablation.
    \item[$F_3$: Preferential effect on violations.] The targeted difference between ungrammatical and grammatical margins is strictly positive and significantly greater than the corresponding random-control difference (one-sided Wilcoxon signed-rank test, $p < \alpha$). This rules out non-selective intervention effects: the ablated features must affect ungrammatical sentences more than grammatical ones.
    \begin{equation}
    \bar{M}_{l,\text{pref}} = \frac{1}{|\mathcal{D}|}\sum_i M_{i,l,\text{pref}}, \ \text{with} \ \bar{M}_{l,\text{pref}} > 0
    \end{equation}
\end{description}
 
A phenomenon is counted as \emph{confirmed} only when $F_1$, $F_2$, and $F_3$ are jointly satisfied. Partial passes (e.g.\ $F_1$ only, $F_1\cap F_2$) are discussed as diagnostic signals of the underlying representation. %rather than full confirmation of violation-specific encoding.
  
\section{Results}
 
\subsection{Cross-Model Overview}

Table~\ref{tab:summary} shows that all models manage to
prefer the grammatical sentences to a comparable degree (74\% to 80\%). 

It also reports, for each model, the number of BLiMP phenomena (out of 13) satisfying partial or complete criteria on at least one layer of the respective model.
\begin{table}[t]
\centering
\small
\setlength{\tabcolsep}{4.8pt}
\caption{Number of phenomena (out of 13) passing each falsification criterion (total) and with significant Wilcoxon test (sig/total); ($p < \alpha$, Bonferroni-corrected $\alpha = 0.00025$ based on the total number of layers); $F_1$: Causal effect on violations, $F_2$: Stability of non-violations; $F_3$: Preferential effect on violations. Combinations require all criteria to hold simultaneously in the same layer.}
\label{tab:summary}
\begin{tabular}{lcccccc}
\toprule
\multirow{2}{*}{\shortstack{Criterion}} & GPT-2 & Gem & Gem & Gem & Qw & Qw \\
& (240M) & (270M) & (1B) & (4B) & (2B) & (9B) \\
\midrule
Acc. & 0.801 & 0.752 & 0.741 & 0.746 &  0.758 & 0.791 \\
\midrule

$F_1$ & 9/11 & 13/13 & 13/13 & 8/8 & 10/10 & 6/6 \\
$F_2$ & 6/6 & 0/4 & 1/2 & 4/4 & 0/1 & 1/2 \\
$F_3$ & 5/10 & 1/1 & 1/1 & 1/1 & 4/8 & 3/9 \\
$F_1 F_2$ & 2/4 & 0/3 & 1/1 & 3/4 & 0/1 & 1/2 \\
$F_2 F_3$ & 2/3 & 0/0 & 0/0 & 0/0 & 0/1 & 0/1 \\
$F_1 F_3$ & 4/6 & 1/1 & 1/1 & 1/1 & 2/4 & 2/2 \\
$F_1 F_2 F_3$ & 1/2 & 0/0 & 0/0 & 0/0 & 0/1 & 0/1 \\

\bottomrule
\end{tabular}
\end{table}
Two patterns hold consistently across models. First, $F_1$ passes broadly: Targeted ablation reliably raises log-likelihood on ungrammatical sentences above the random control baseline, confirming that high-$\Phi$ features are correlated to constraint-violating inputs. Second, $F_2$ fails in most cases, indicating that the top-$k$ features selected by $\Phi$ also causally affect well-formed sentences. Third, $F_3$ also fails frequently: even when ablation effects are present, they are often comparable in magnitude across grammatical and ungrammatical conditions, suggesting the identified features affect sentence likelihood broadly rather than selectively targeting violations.
\begin{table*}[t]
\centering
\small
\caption{Per-phenomenon ablation results at best layer $l^*$ (GPT-2, top-$k=15$), re-evaluated using layer-wise selection by number of passed criteria. * denotes statistical significance under a Wilcoxon test ($p < \alpha$), with Bonferroni-corrected threshold $\alpha = 0.00025$ based on the total number of layers.}
\label{tab:detail_recomputed}
\begin{tabular}{lrrrrrrrrccc}
\toprule
Phenomenon & $l^*$      
           & $\bar{M}_{l,\text{bad}}$ & $\bar{M}_{l,\text{bad}}^{\text{ctrl}}$
           & $\bar{M}_{l,\text{good}}$ & $\bar{M}_{l,\text{good}}^{\text{ctrl}}$
           & $\bar{M}_{l,\text{pref}}$ & $\bar{M}_{l,\text{pref}}^{\text{ctrl}}$
           & $F_1$ & $F_2$ & $F_3$ \\
\midrule

Subject--verb agreement  & L10 & $+0.2627^{*}$ & $-0.0210$ & $+0.2291^*$ & $-0.0220$ & $-0.1209$ & $-0.1534$ & \checkmark & \texttimes & \texttimes \\

Det.--noun agreement     & L10 & $+0.1852^*$ & $+0.0251$ & $+0.0457^*$ & $+0.0179$ & $-0.5210^*$ & $-0.6534$ & \checkmark & \texttimes & \texttimes \\

Anaphor agreement        & L10 & $-0.1702$ & $-0.0178$ & $-0.2109^*$ & $-0.0174$ & $-3.7746$ & $-3.8157$ & \texttimes & \texttimes & \texttimes \\

Filler-gap dependency    & L10 & $+0.5944^*$ & $+0.0477$ & $+0.2036$ & $+0.0374$ & $+0.3068^*$ & $-0.0736$ & \checkmark & \texttimes & \checkmark \\

Island effects           & L10 & $-0.7963$ & $-0.0214$ & $-0.7346$ & $-0.0116$ & $+0.0038$ & $+0.0557$ & \texttimes & \texttimes & \checkmark \\

NPI licensing            & L7  & $+0.2465$ & $+0.0129$ & $+0.1702^*$ & $+0.0123$ & $+3.8955$ & $+3.8198$ & \checkmark & \texttimes & \checkmark \\

Argument structure       & L10 & $-0.0188$ & $+0.0447$ & $-0.0252^*$ & $+0.0443$ & $+0.1279$ & $+0.1219$ & \texttimes & \texttimes & \checkmark \\

S-selection              & L10 & $+0.4257^*$ & $+0.0005$ & $+0.2156^*$ & $+0.0058$ & $+0.9772^*$ & $+0.7618$ & \checkmark & \texttimes & \checkmark \\

Control / raising        & L10 & $+0.0221^*$ & $-0.0664$ & $-0.0880$ & $-0.0728$ & $+3.1758^*$ & $+3.0721$ & \checkmark & \texttimes & \checkmark \\

Binding                  & L7  & $+0.0941^*$ & $+0.0275$ & $+0.0831$ & $+0.0249$ & $-1.0357^*$ & $+0.0604$ & \checkmark & \texttimes & \texttimes \\

Ellipsis                 & L7  & $+0.3789^*$ & $+0.0306$ & $+0.2942^*$ & $+0.0221$ & $-3.3554$ & $-3.4316$ & \checkmark & \texttimes & \texttimes \\

Quantifiers              & L7  & $+0.4390^*$ & $-0.0515$ & $+0.1290^*$ & $-0.0495$ & $+0.5295^*$ & $+0.2175$ & \checkmark & \texttimes & \checkmark \\

Irregular forms          & L7  & $+0.0843$ & $+0.0330$ & $+0.0156^*$ & $+0.0286$ & $+0.7549^*$ & $+0.6905$ & \checkmark & \texttimes & \checkmark \\

\bottomrule
\end{tabular}
\end{table*}
The joint criterion $F_1 \cap F_2 \cap F_3$ is satisfied by one phenomenon on GPT-2 model, even when $F_1$ is widely met. This gap suggests that causal sensitivity alone is not sufficient evidence for violation-specific encoding. True selectivity requires both stability on well-formed inputs and stronger effects on ungrammatical ones, but these conditions rarely co-occur under the current ablation framework. 

Taken together, the overarching hypothesis is not supported: while models contain information relevant to grammatical violations, this information is not cleanly separable from general sentence-level representations under the current intervention framework.

\FloatBarrier
\subsection{Per-Phenomenon Analysis (GPT-2 Reference)}
Table~\ref{tab:detail_recomputed} reports per-phenomenon ablation results under a revised layer-selection rule. For each phenomenon, the selected layer is the one that maximizes the number of satisfied criteria. This selection strategy differs from Table~\ref{tab:summary}, which leads to the observed differences in aggregate counts.

There are two main observational findings. (1) The best-performing layers are concentrated in the middle of the network, consistent with prior work \citep{petersDissecting2018, li2021bert}, suggesting a greater concentration of complex representations in intermediate layers. (2) $F_1$ and $F_3$ often co-occur, as both can be driven by the same sensitivity signal, while $F_2$ is rarely satisfied. Although $F_2$ may hold in other layers, it is primarily disrupted by interference from features that also affect well-formed sentences. However, we also tested the top-$1$ setting, and $F_2$ still remains largely unfulfilled. This suggests that the issue is not solely due to feature aggregation, but rather reflects a more intrinsic entanglement of features that also affect well-formed sentences. This is consistent with prior observations of the feature absorption effect in SAEs \citep{chanin2026absorption}.

\section{Limitation}
Our work is subject to two limitations. First, computational constraints restrict the analysis to a limited range of model sizes, which may affect the generality of the findings to larger or more recent LLMs. Second, our evaluation is based on a specific set of falsification tests, which does not fully capture the breadth of linguistic phenomena relevant to constraint violation processing; in particular, alternative falsification frameworks targeting surprisal, semantic anomalies, and plausibility could provide complementary perspectives. These limitations are nonetheless not critical for the interpretation of our results, since the main finding is negative.

\section{Conclusion}
This work investigated whether language models encode dedicated latent features representing linguistic constraint violations. Using targeted activation ablation and a conjunctive falsification framework, we tested whether candidate features selectively affect ungrammatical sentences while preserving grammatical ones. Across six language models and thirteen BLiMP phenomena, support for the hypothesis was limited: although many phenomena satisfied individual criteria, very few satisfied all three simultaneously.

Overall, this study yields negative results in two respects: (1) the falsification criteria are not jointly satisfied across phenomena, and (2) no features are consistently shared across all linguistic categories. We hypothesize that the negative results may be due to: (1) the sensitivity score failing to uncover features that satisfy $F_2$ and $F_3$; (2) sparse autoencoders not recovering features corresponding to $F_2$ and $F_3$, which may instead be absorbed into other features; and (3) the relevant features being distributed across multiple layers rather than localized to a single layer. 
\FloatBarrier
\section*{Impact Statement}
This paper presents work whose goal is to advance our understanding of the fundamental principles of language processing in large language models. It does not have direct ethical ramifications above and beyond the general concerns of the field.


\bibliography{example_paper}
\bibliographystyle{icml2026}

%%%%%%%%%%%%%%%%%%%%%%%%%%%%%%%%%%%%%%%%%%%%%%%%%%%%%%%%%%%%%%%%%%%%%%%%%%%%%%%
%%%%%%%%%%%%%%%%%%%%%%%%%%%%%%%%%%%%%%%%%%%%%%%%%%%%%%%%%%%%%%%%%%%%%%%%%%%%%%%
% APPENDIX
%%%%%%%%%%%%%%%%%%%%%%%%%%%%%%%%%%%%%%%%%%%%%%%%%%%%%%%%%%%%%%%%%%%%%%%%%%%%%%%
%%%%%%%%%%%%%%%%%%%%%%%%%%%%%%%%%%%%%%%%%%%%%%%%%%%%%%%%%%%%%%%%%%%%%%%%%%%%%%%
% \newpage
% \appendix
% \onecolumn
%%%%%%%%%%%%%%%%%%%%%%%%%%%%%%%%%%%%%%%%%%%%%%%%%%%%%%%%%%%%%%%%%%%%%%%%%%%%%%%
%%%%%%%%%%%%%%%%%%%%%%%%%%%%%%%%%%%%%%%%%%%%%%%%%%%%%%%%%%%%%%%%%%%%%%%%%%%%%%%

\end{document}